\documentclass[conference]{IEEEtran}
\IEEEoverridecommandlockouts
\usepackage{cite}
\usepackage{amsmath,amssymb,amsfonts}
\usepackage{algorithmic}
\usepackage{graphicx}
\usepackage{textcomp}
\usepackage{xcolor}
\usepackage[bookmarks=false]{hyperref}

\usepackage[group-separator={,}]{siunitx}
\def\BibTeX{{\rm B\kern-.05em{\sc i\kern-.025em b}\kern-.08em
    T\kern-.1667em\lower.7ex\hbox{E}\kern-.125emX}}

\usepackage{booktabs} %top/mid/bottom rule

\definecolor{red}{RGB}{255,0,0}
\definecolor{green}{RGB}{0, 102, 0}
\definecolor{maron}{RGB}{102,51,0}
\definecolor{orange}{RGB}{255,128,0}
\definecolor{blue}{RGB}{0,153,153}

\begin{document}

\title{Leveraging Subword Embeddings for Multinational Address Parsing}

\author{\IEEEauthorblockN{Marouane Yassine, David Beauchemin, François Laviolette, Luc Lamontagne}
\IEEEauthorblockA{\textit{Department of Computer Science and Software Engineering, Laval University} \\
\textit{Group for Research in Artificial Intelligence of Laval University (GRAIL)}\\
Québec, Canada \\
marouane.yassine.1@ulaval.ca, david.beauchemin.5@ulaval.ca, francois.laviolette@ift.ulaval.ca, luc.lamontagne@ift.ulaval.ca}
}

\maketitle

\begin{abstract}
Address parsing consists of identifying the segments that make up an address such as a street name or a postal code. Because of its importance for tasks like record linkage, address parsing has been approached with many techniques. Neural network methods defined a new state-of-the-art for address parsing. While this approach yielded notable results, previous work has only focused on applying neural networks to achieve address parsing of addresses from one source country. We propose an approach in which we employ subword embeddings and a Recurrent Neural Network architecture to build a single model capable of learning to parse addresses from multiple countries at the same time while taking into account the difference in languages and address formatting systems. We achieved accuracies around $99~\%$ on the countries used for training with no pre-processing nor post-processing needed. We explore the possibility of transferring the address parsing knowledge obtained by training on some countries' addresses to others with no further training in a zero-shot transfer learning setting. We achieve good results for $80~\%$ of the countries (33 out of 41), almost $50~\%$ of which (20 out of 41) is near state-of-the-art performance. In addition, we propose an open-source Python implementation of our trained models\footnote{https://github.com/GRAAL-Research/deepparse}.
\end{abstract}


\section{Introduction}
\emph{Address Parsing} is the task of decomposing an address into the different components it is made of. This task is an essential part of many applications, such as geocoding and record linkage. Indeed, to find a particular location based on textual data, it is quite useful to detect the different parts of an address  to make an informed decision. Similarly, comparing two addresses to decide whether two or more database entries refer to the same entity can prove to be quite difficult and prone to errors if based on methods such as edit distance algorithms given the various address writing standards.

There have been many efforts to solve the address parsing problem. From rule-based techniques \cite{rule-based} to probabilistic approaches and neural network models \cite{8615844}, a lot of progress has been made in reaching an accurate segmentation of addresses. These previous pieces of work did a remarkable job at finding solutions for the challenges related to the address parsing task. However, most of these approaches either do not take into account parsing addresses from different countries or do so but at the cost of a considerable amount of meta-data and substantial data pre-processing pipelines \cite{rnn-parsing, hmm-parsing, crf-parsing, feedforward-parsing}.

Our work comes with two objectives. Firstly, we propose an approach for multinational address parsing using a Recurrent Neural Network (RNN) architecture. We start by addressing the multilingual aspect of the problem by employing multilingual sub-word units. Then we train an architecture composed of an embedding layer followed by a sequence-to-sequence (Seq2Seq) model. Secondly, we evaluate the degree to which a model trained on countries' addresses data can perform well at parsing addresses from other countries.

\section{Related work}

Since address parsing is a sequence tagging task, it has been tackled using probabilistic methods mainly based on Hidden Markov Models (HMM) and Conditional Random Fields (CRF) \cite{hmm-parsing, crf-parsing, 8615844}. For instance, \cite{hmm-parsing} proposed a large scale HMM-based parsing technique capable of segmenting a large number of addresses, whilst being robust to possible irregularities in the input data. In addition, \cite{crf-parsing} implemented a discriminative model using a linear-chain CRF coupled with a learned Stochastic Regular Grammar (SRG). This approach allowed the authors to better address the complexity of the features while capturing higher-level dependencies by applying the SRG on the CRF outputs as a score function, thus taking into account the possible lack of features for a particular token in a lexicon-based model. These probabilistic methods usually rely on structured data as well as some sort of prior knowledge of this data for feature extraction or in order to implement algorithms such as Viterbi \cite{viterbi}, especially in the case of generative methods. 

In recent years, new methods \cite{rnn-parsing, 8615844} utilizing the power of neural networks have been proposed as solutions for the address parsing problem. Using a single hidden layer feed-forward model, \cite{feedforward-parsing} achieved state-of-the-art performance. Their approach, however, relied on a pipeline of pre-processing and post-processing so as to deal with the different structures of address writing, as well as the possible prediction errors. For instance, the input data is normalized to reduce noise and to standardize the many variations that can refer to the same word, such as \emph{road} and \emph{rd}. In addition, the model's predictions are put through a rule-based validation step to make sure that they fit known patterns. In contrast, \cite{rnn-parsing} proposed a deep learning approach based on the use of RNN. Their experiments focused on comparing the performance of both unidirectional and bidirectional vanilla RNN and Long-Short Term Memory Models (LSTM) \cite{LSTM}, as well as a Seq2Seq. The models achieved high accuracy on test sets with the Seq2Seq leading the scoreboard on most of them with no particular pre-processing needed during the inference process.

Note however that despite reaching notable performances, the aforementioned approaches are limited to parsing addresses from a single country and would need to be adjusted to support a multinational scope of address parsing. To tackle this problem, Libpostal\footnote{https://github.com/openvenues/libpostal}, a library for international address parsing, has been proposed. This library uses a CRF-based model trained with an averaged Perceptron for scalability. The model was trained on data from each country in the world and was able to achieve a $99.45~\%$ full parse accuracy\footnote{The accuracy was computed considering the entire sequence and was not focused on individual tokens.}. However, this requires putting addresses through a heavy pre-processing pipeline before feeding them to the prediction model. It is our understanding that no neural network approaches were proposed for multinational address parsing with a single model. This work aims at building a single model solution capable of parsing addresses from multiple countries, as well as exploring the possibility of zero-shot transfer from some countries addresses to others'.

\section{Subword embeddings}

The use of subword embeddings has become popular across Natural Language Processing tasks given the performance enhancements they provide to neural network models. Word embeddings \cite{word2vec, glove} are usually augmented by character-level or subword-level information before being fed to the model as inputs, thus granting it a more meaningful representation of words. This strategy is employed by the word embeddings library fastText \cite{fasttext} in which a representation of words as character n-grams is used along with words representations in order to produce embeddings. This approach allows for a model capable of producing richer embeddings, as well as embeddings for out-of-vocabulary words (OOV), which are computed as the sum of their n-gram fractions' embeddings. For example, the embedding of the OOV word "H1A 1B1" using a bi-gram is the sum of the fractions' embedding of \{H1, 1A, A1, 1B, B1\}. 

\subsection{Byte-pair Encoding}
Byte-pair encoding (BPE) \cite{bpe} is a data compression algorithm which iteratively replaces the most frequent occurrences of adjacent bytes with a new set of bytes to find a more compact representation of the said data. A new approach for word segmentation based on the BPE algorithm was introduced by \cite{sennrich-etal-2016-neural}. Their technique, which was proposed to solve the OOV problem in Neural Machine Translation (NMT), consists of representing text as a sequence of characters that are iteratively merged using the same reasoning behind BPE. This approach paved the way for the authors to address NMT with an open-vocabulary solution. Another application of BPE is BPEmb \cite{bpemb}, a set of embedding models that were trained to produce subword embeddings based on a BPE decomposition of text. BPEmb offers pre-trained models on 275 languages, as well as MultiBPEmb, which is a single model trained on the shared vocabulary of the 275 languages. These models were shown to have a performance similar to other subword embedding techniques on an entity typing task while outperforming these techniques on some languages.

\section{Architecture}
\begin{figure*}[!h]
    \centering
    \includegraphics[width=\linewidth]{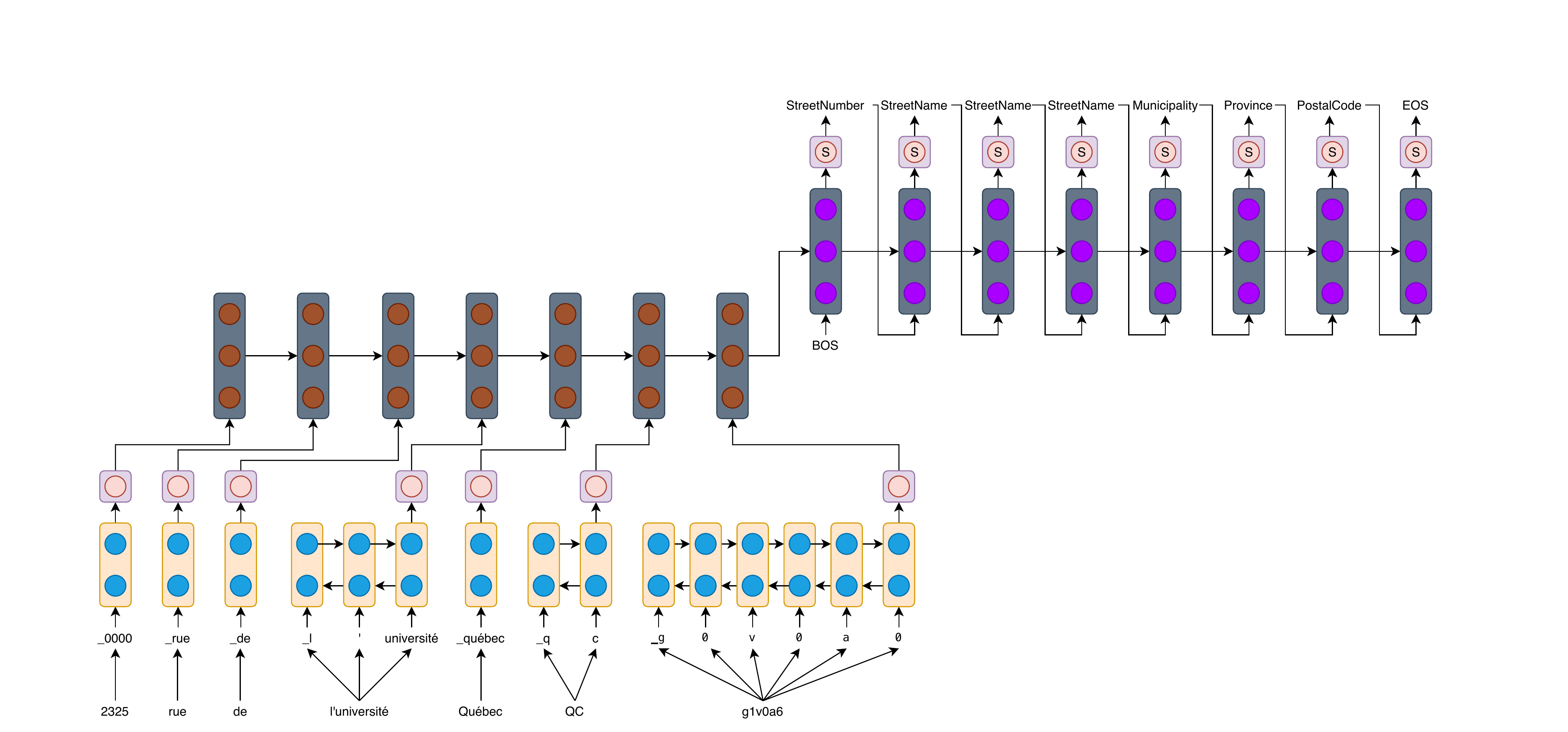}
    \caption{Illustration of our architecture using the BPEmb embedding model. Each word in the address is encoded using MultiBPEmb (the BPE segmentation algorithm replaces the numbers in the address by zeros). The subword embeddings are fed to the BiLSTM (rounded rectangle with two circles). The last hidden state for each word is run through a fully connected layer (rounded rectangle with one circle). The resulting embeddings are given as input to the Seq2Seq (rounded rectangle with three circles). The 'S' in the fully connected layer following the Seq2Seq decoder stands for the Softmax function.}
    \label{fig:my_label}
\end{figure*}

The following section describes the architecture of our model, which is composed of an embedding model (\autoref{subsec:embedding}) and a tagging model (\autoref{subsec:tagging}) as shown in \autoref{fig:my_label}.

\subsection{Embedding Model}
\label{subsec:embedding}
Since our main objective is to build a single neural network for parsing addresses from multiple countries, it is necessary to have access to embeddings for different languages at runtime. Some libraries, such as fastText \cite{fasttext-aligned} and MUSE \cite{muse}, offer alignment vectors that enable the projection of word embeddings from different languages in the same space. However, these techniques would require detecting the source language as well as specifying the target language to use the proper alignments, which we consider an unnecessary overhead for the task at hand. 
To resolve the embedding issue, we propose the following two methods.

First, we use a fixed pre-trained monolingual fastText model (pre-trained on the French language) (\textbf{fastText}). We chose French embeddings since the French language shares Latin roots with many languages in our test set. It is also due to the considerable size of the corpus on which these embeddings were trained.

Second, we use an encoding of words using MultiBPEmb and merge the obtained embeddings for each word into one word embedding using a RNN. This method has been shown to give good results in a multilingual setting \cite{heinzerling-strube-2019-sequence}.  Our RNN network of choice is a Bidirectional LSTM (Bi-LSTM) with a hidden state dimension of 300. We build the word embeddings by running the concatenated forward and backward hidden states corresponding to the last time step for each word decomposition through a fully connected layer of which the number of neurons is equal to the dimension of the hidden states. This approach produces a 300-dimensional word embeddings. We refer to this embeddings model technique as \textbf{BPEmb}.

We run a comparison of the two methods (\textbf{fastText} and \textbf{BPEmb}) to evaluate which one gives better results in our setting.

\subsection{Tagging Model}
\label{subsec:tagging}

Our downstream tagging model is a Seq2Seq model consisting of a one-layer unidirectional LSTM encoder and a one-layer unidirectional LSTM decoder followed by a fully-connected linear layer with a softmax activation. Both the encoder's and decoder's hidden states are of dimension $1024$. The embedded address sequence is fed to the encoder that produces hidden states, the last of which is used as a context vector to initialize the decoder's hidden states. The decoder is then given a Beginning Of Sequence (BOS) token as input, and at each time step, the prediction from the last step is used as input. To better adapt the model to the task in hand and to facilitate the convergence process, we only require the decoder to produce a sequence with the same length as the input address. This approach differs from the traditional Seq2Seq architecture in which the decoder makes predictions until it predicts the EOS token. The decoder's outputs are forwarded to the linear layer of which the number of neurons is equal to the tag space dimensionality. The softmax activation function computes probabilities over the linear layer's outputs to predict the most likely token at each time step.

\section{Data}

Our dataset was built using the open-source data on which Libpostal's models were trained and of which we have collected the address data of 61 countries. Twenty countries were used for multinational training with a sample size of \num{100000} addresses per country while the rest of the samples was left out as holdout for testing. The other countries' data was also left for zero-shot transfer evaluation. Tables \ref{tab:number_holdout_samples} and \ref{tab:number_zeroshot_samples} show the number of samples per country in both test sets ordered by number of examples per country. The color in the table will be discussed later on.

\begin{figure}[h!]
    \centering
    \includegraphics[width=9cm]{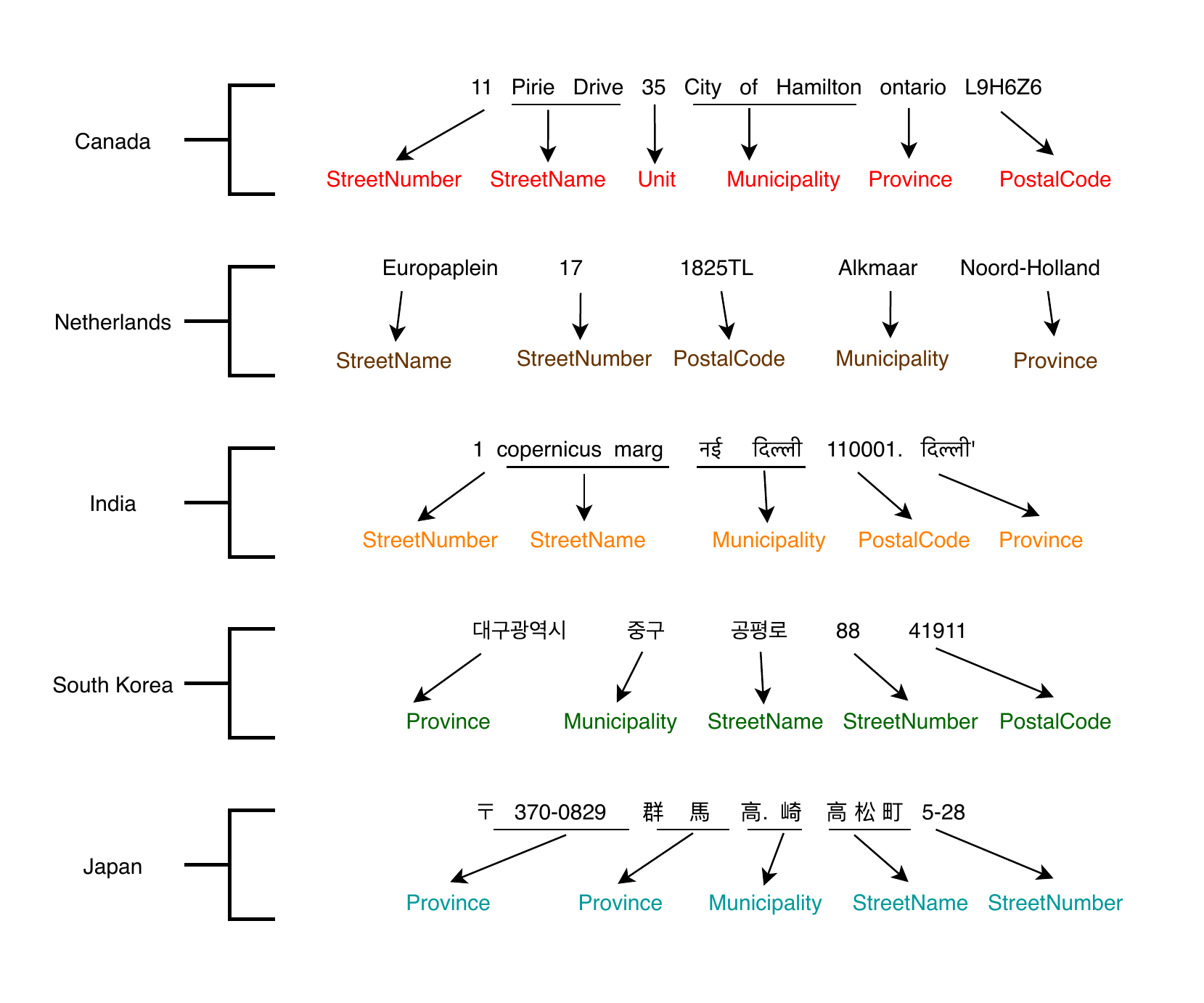}
    \caption{Address samples per country}
    \label{fig:address_examples}
\end{figure}

We introduce eight tags, namely StreetNumber, StreetName, Unit, Municipality, Province, PostalCode, Orientation, and GeneralDelivery, as opposed to Libpostal, which utilizes 20 tags. This was motivated by the common presence of the chosen tags in most of the countries that are included in our datasets. Also, it is not guaranteed that all addresses contain each tag category's elements since some addresses might not contain elements of some tag categories. \autoref{fig:address_examples} shows address samples for different countries with the corresponding tags. Each color represents one of the five different patterns present in our dataset \cite{rhind1998global}. We also find that some countries' address format is composed of different patterns (e. g. Belarus that use the second and fifth patterns). No color is used for these countries.

\begin{table*}[h!]
    \centering
    \caption{Number of samples per country in the holdout test set for training countries} \label{tab:number_holdout_samples}
    \begin{tabular}{cccccccc}
    \toprule
        Country & Number of samples & Country & Number of samples & Country & Number of samples & Country & Number of samples \\ \midrule
        \color{red}United States & \num{8000000} & \color{maron}Germany & \num{1576059} & \color{maron}Poland & \num{459522} &  \color{maron}Czechia & \num{195269} \\ 
        \color{red}Brazil & \num{8000000} & \color{maron}Spain & \num{1395758} & \color{maron}Norway & \num{405649} & \color{maron} Italy & \num{178848} \\ 
        \color{green}South Korea & \num{6048106} & \color{maron}Netherlands & \num{1202173} & \color{maron}Austria & \num{335800} &  \color{maron}France & \num{20050} \\ 
        \color{red}Australia & \num{5428043} & \color{red}Canada & \num{910891} & \color{maron}Finland & \num{280219} &  \color{red}United Kingdom & \num{14338} \\ 
        \color{maron}Mexico & \num{4853349} & \color{maron}Switzerland & \num{474240} & \color{maron}Denmark & \num{199694} &  \color{red}Russia & \num{8115} \\
         \bottomrule
    \end{tabular}
\end{table*}

\begin{table*}[h!]
    \centering
    \caption{Number of samples per country in the zero-shot test set} \label{tab:number_zeroshot_samples}
    \begin{tabular}{cccccccc}
    \toprule
        Country & Number of samples & Country & Number of samples & Country & Number of samples & Country & Number of samples \\ \midrule
        \color{maron}Belgium & \num{66182} & \color{maron}Slovenia & \num{9773} & \color{maron}Réunion & \num{2514} &  \color{red}Bangladesh & \num{888} \\ 
    \color{maron}Sweden & \num{32291} & \color{red}Ukraine & \num{9554} & \color{maron}Moldova & \num{2376} &  \color{maron}Paraguay & \num{839} \\ 
    \color{maron}Argentina & \num{27692} & Belarus & \num{7590} & \color{red}Indonesia & \num{2259} &  \color{maron}Bosnia & \num{681} \\ 
    \color{orange}India & \num{26084} & \color{maron}Serbia & \num{6792} & \color{red}Bermuda & \num{2065} &  \color{maron}Cyprus & \num{836} \\ 
    \color{maron}Romania & \num{19420} & \color{maron}Croatia & \num{5671} & \color{maron}Malaysia & \num{2043} &   \color{red}Ireland & \num{638} \\ 
    \color{maron}Slovakia & \num{18975} & \color{maron}Greece & \num{4974} & \color{red}South Africa & \num{1388} &  \color{maron}Algeria & \num{601} \\ 
    Hungary & \num{17460} & \color{red}New Zealand & \num{4678} & \color{red}Latvia & \num{1325} &  \color{maron}Colombia & \num{569} \\ 
    \color{blue}Japan & \num{14089} & \color{maron}Portugal & \num{4637} & \color{blue}Kazakhstan & \num{1087} &  Uzbekistan & \num{505} \\ 
    \color{maron}Iceland & \num{13617} & \color{maron}Bulgaria & \num{3716} & \color{maron}New Caledonia & \num{1036} &  & \\ 
    \color{red}Venezuela & \num{10696} & \color{maron}Lithuania & \num{3126} & \color{maron}Estonia & \num{1024} &  &  \\ 
    Philippines & \num{10471} & \color{maron}Faroe Islands & \num{2982} 
    & \color{red}Singapore & \num{968} & & \\
         \bottomrule
    \end{tabular}
\end{table*}

\section{Experiments}
\label{sec:exp}
For our experiments, we trained our two models (\textbf{fastText} and \textbf{BPEmb}) five times each\footnote{Using each of the following seeds $\{5, 10, 15, 20, 25\}$. When a model didn't converge (a high loss value on train and validation), we retrain the model using a different seed (30).} for 200 epochs with a batch size of \num{2048}. An early stopping with a patience of fifteen epochs was also applied during training. We initialize the learning rate at 0.1 and use learning rate scheduling to lower it by a factor of 0.1 after ten epochs without loss reduction. Our loss function of choice is the Cross-Entropy loss due to its suitability for the softmax function. The optimization is done through Stochastic Gradient Descent.

Also, to speed up the convergence, we use teacher forcing \cite{6795228}, a method that consists of using the ground truth instead of the previous time step's prediction as input for the decoder during training. We do so by randomly sampling part of the training data at runtime. $80~\%$ of the training datasets were used to train the models, and $20~\%$ was kept for validation. The architecture, as well as the training of the models, were implemented using Pytorch \cite{pytorch} and Poutyne \cite{poutyne}.

\subsection{Evaluation Procedure}

We train our two models on our multinational dataset, the difference between the models being the word embedding method employed (\textbf{fastText} and \textbf{BPEmb}). Each model has been trained five times and we report the models' mean accuracy and standard deviation on the per-country holdout data in \autoref{table:multinational-holdout}. The accuracy for each sequence is computed as the proportion of the tags predicted correctly by the model. As such, predicting all the tags for a sequence correctly yields a perfect accuracy. More precisely, errors in tag predictions have an impact on the accuracy for a given sequence. However, the accuracy will not be null unless all the predicted tags for the sequence are incorrect. Our two models mean results are compared to those of Libpostal evaluated on the same validation dataset. However, since Libpostal’s preprocessing step has introduced an incompatibility between the input data and the ground truth, we were obliged to discard a few addresses (less than $0.3~\%$ of them) during the scores’ generation for some countries (South Korea, Germany, Norway, Poland, Serbia, Japon and Bulgaria).

\begin{table*}[h!]
    \centering
    \caption{Multinational models' mean accuracy (and standard deviation) on holdout datasets for training countries}
    \label{table:multinational-holdout}
    \vspace{6pt}
    \begin{tabular}{cccccccc}
    \toprule
    Country & Libpostal & FastText & BPEmb & Country & Libpostal & FastText & BPEmb\\
    \midrule
    United States & $\mathbf{99.95}$ & $99.61 \pm 0.09$ & $99.67 \pm 0.09$ & Poland & $\mathbf{99.93}$ & $99.69 \pm 0.07$ & $99.89 \pm 0.04$\\
    Brazil & $\mathbf{99.94}$ & $99.40 \pm 0.10$ & $99.42 \pm 0.15$ & Norway & $\mathbf{99.96}$ & $99.46 \pm 0.06$ & $98.41 \pm 0.63$\\
    South Korea & $77.25$ & $99.96 \pm 0.01$ & $\mathbf{100.00 \pm 0.00}$ & Austria & $\mathbf{99.38}$ & $99.28 \pm 0.03$ & $98.98 \pm 0.22$\\
    Australia & $98.86$ & $99.68 \pm 0.05$ & $\mathbf{99.80 \pm 0.05}$ & Finland & $99.84$ & $99.77 \pm 0.03$ & $\mathbf{99.87 \pm 0.01}$\\
    Mexico & $99.62$ & $99.60 \pm 0.06$ & $\mathbf{99.68 \pm 0.06}$ & Denmark & $\mathbf{99.96}$ & $99.71 \pm 0.07$ & $99.90 \pm 0.03$\\
    Germany & $99.82$ & $99.77 \pm 0.04$ & $\mathbf{99.89 \pm 0.03}$ & Czechia & $99.83$ & $99.57 \pm 0.09$ & $\mathbf{99.89 \pm 0.04}$\\
    Spain & $\mathbf{99.89}$ & $99.75 \pm 0.05$ & $99.85 \pm 0.04$ & Italy & $99.66$ & $99.73 \pm 0.05$ & $\mathbf{99.81 \pm 0.05}$\\
    Netherlands & $\mathbf{99.96}$ & $99.61 \pm 0.07$ & $99.88 \pm 0.03$ & France & $\mathbf{99.84}$ & $99.66 \pm 0.08$ & $99.69 \pm 0.11$\\
    Canada & $\mathbf{99.94}$ & $99.79 \pm 0.05$ & $99.87 \pm 0.04$ & United Kingdom & $99.28$ & $99.61 \pm 0.10$ & $\mathbf{99.74 \pm 0.08}$\\
    Switzerland & $99.58$ & $99.53 \pm 0.09$ & $\mathbf{99.75 \pm 0.08}$ & Russia & $\mathbf{99.95}$ & $99.03 \pm 0.24$ & $99.67 \pm 0.11$\\
    \bottomrule
    \end{tabular}
\end{table*}

\begin{table*}[h!]
    \centering
    \caption{Multinational models' z-test significance test on holdout datasets for training countries (\textbf{bold} value are rejected null hypothesis with $\alpha = 0.001$)}
    \label{table:multinational-holdout-stats}
    \vspace{6pt}
    \begin{tabular}{cccccccc}
        \toprule
        Country &  \begin{tabular}[c]{@{}c@{}}Libpostal\\ FastText\end{tabular} & \begin{tabular}[c]{@{}c@{}}Libpostal\\ BPEmb\end{tabular} & \begin{tabular}[c]{@{}c@{}}FastText\\ BPEmb\end{tabular} & Country & \begin{tabular}[c]{@{}c@{}}Libpostal\\ FastText\end{tabular} & \begin{tabular}[c]{@{}c@{}}Libpostal\\ BPEmb\end{tabular} & \begin{tabular}[c]{@{}c@{}}FastText\\ BPEmb\end{tabular}\\
        \midrule
        United States & 144.58 & 127.52 & -20.56 & Poland & 26.86 & 7.19 & -20.73\\
        Brazil & 185.68 & 180.75 & -6.12 & Norway & 41.96 & 77.70 & 46.01\\
        South Korea & -1243.10 & -1246.13 & -48.29 & Austria & 5.31 & 18.33 & 13.11\\
        Australia & -157.17 & -190.27 & -41.23 & Finland & 5.46 & \textbf{-3.15} & -8.56\\
        Mexico & 5.36 & -17.76 & -23.10 & Denmark & 19.02 & 6.49 & -13.70\\
        Germany & 9.04 & -17.43 & -26.18 & Czechia & 14.87 & -5.27 & -19.50\\
        Spain & 27.96 & 10.05 & -18.32 & Italy & -3.51 & -8.28 & -4.83\\
        Netherlands & 58.93 & 21.80 & -42.13 & France & 3.54 & \textbf{3.11} & \textbf{-0.45}\\
        Canada & 28.50 & 16.08 & -13.38 & United Kingdom & -3.72 & -5.55 & \textbf{-1.95}\\
        Switzerland & 3.85 & -14.05 & -17.80 & Russia & 8.17 & 3.99 & -5.10\\
        \bottomrule
        \end{tabular}
\end{table*}

\subsection{Multinational Evaluation}
First, we find that \textbf{BPEmb} gives better results than Libpostal for 9 of 20 countries, even if we have trained on just \num{100000} examples of each country compared to near millions per country for Lipbostal. These results show that our approach can achieve similar results whilst requiring less data and without the need of pre-processing nor post-processing. 

Second, we find the model using BPEmb embeddings to have the best performance across the board without considering Libpostal. The \textbf{BPEmb} model achieves better results than \textbf{fastText} in most cases (all except Norway and Austria). We find that South Korea is the only country where a perfect accuracy was achieved using \textbf{BPEmb} (for four seeds out of five). Since South Korea is the only country using a different pattern in the training set where the province and municipality occur before the street name, it seems that our models might have memorized this particular pattern. To validate this intuition we randomly reordered \num{6000} South Korean addresses to follow either the first (red) or the second (brown) address pattern (equally divided between the two). We observe, after this reordering, that the mean accuracy drops to $28.04~\%$ considering that using a random tags annotation, we get a $12.29~\%$ accuracy.

It was also interesting to notice the model accuracy is good when using fastText monolingual word embeddings, especially on South Korean addresses despite the entirely different alphabet. These results illustrate that our model, regardless of the embeddings model, learned the representation of an address sequence even if the words' representations are not native to the language (French vs Korean). 

Overall, all our models achieve state-of-the-art performance on our dataset while using less data than previous approaches. However, \textbf{BPEmb} generates better results than \textbf{fastText} and similar ones to Libpostal. To further assess' models' performance, we report the models' z-test significance test in \autoref{table:multinational-holdout-stats}. Our z-test null hypothesis is that the pair of models have equal performances, meaning that values smaller or greater than $|3.290527|$ allow us to reject the null hypothesis with $\alpha = 0.001$. A positive value means that the first model (left) has a significantly better performance than the second (right), and a negative value means the opposite. We observe that \textbf{BPEmb} has a significantly better performance for 8 out of 20 countries compared to Libpostal, which means that the performance between the two is quite similar. Besides, we observe that \textbf{BPEmb} has almost always significantly better performance than \textbf{fastText}. Considering these results, we can conclude that \textbf{BPEmb} is better than \textbf{fastText} and \textbf{BPEmb} is similar to Libpostal but using less training data and no pre-processing nor post-processing.

\subsection{Zero-shot Evaluation}
\label{sec:zeroshot}

Since training a deep learning model to parse addresses from every country in the world would require a significant amount of data and resources, our ongoing work aims at achieving domain adaptation to be able to train on a reasonable amount of data and generalize to data from different sources. We begin by exploring how well our architecture can generalize in a zero-shot manner. To this end, we test each of our five trained models using the two embedding settings on address data from countries not seen during the training. The results are reported in \autoref{table:zero-shot-results} ordered by dataset size. We choose not to bold Libpostal scores when they are the best since their model was trained on these addresses, making the comparison unfair to our models. 

\begin{table*}[h!]
    \centering
    \caption{Zero-shot transfer models' mean accuracy (and standard deviation) per country}
    \label{table:zero-shot-results}
    \vspace{6pt}
        \begin{tabular}{cccccccc}
            \toprule
            Country & Libpostal & FastText & BPEmb & Country & Libpostal & FastText & BPEmb\\
            \midrule
            Belgium & 99.77 & $\mathbf{88.14 \pm 1.04}$ & $87.29 \pm 1.40$ & Faroe Islands & 99.99 & $74.14 \pm 1.83$ & $\mathbf{85.50 \pm 0.11}$\\
            Sweden & 99.96 & $81.59 \pm 4.53$ & $\mathbf{90.76 \pm 3.03}$ & Réunion & 89.15 & $\mathbf{96.80 \pm 0.45}$ & $93.67 \pm 0.26$\\
            Argentina & 99.34 & $86.26 \pm 0.47$ & $\mathbf{88.04 \pm 0.83}$ & Moldova & 99.76 & $\mathbf{90.18 \pm 0.79}$ & $86.89 \pm 3.01$\\
            India & 97.27 & $69.09 \pm 1.74$ & $\mathbf{80.04 \pm 3.24}$ & Indonesia & 98.40 & $64.31 \pm 0.84$ & $\mathbf{70.28 \pm 1.64}$\\
            Romania & 99.85 & $\mathbf{94.49 \pm 1.52}$ & $91.65 \pm 1.21$ & Bermuda & 98.94 & $92.31 \pm 0.60$ & $\mathbf{93.70 \pm 0.35}$\\
            Slovakia & 99.81 & $82.10 \pm 0.98$ & $\mathbf{90.31 \pm 3.88}$ & Malaysia & 97.98 & $78.93 \pm 3.78$ & $\mathbf{94.16 \pm 0.49}$\\
            Hungary & 99.69 & $\mathbf{48.92 \pm 3.59}$ & $25.51 \pm 2.60$ & South Africa & 99.84 & $95.31 \pm 1.68$ & $\mathbf{96.87 \pm 0.96}$\\
            Japan & 92.92 & $\mathbf{41.41 \pm 3.21}$ & $35.33 \pm 1.28$ & Latvia & 99.36 & $\mathbf{93.66 \pm 0.64}$ & $74.78 \pm 4.33$\\
            Iceland & 99.84 & $96.55 \pm 1.20$ & $\mathbf{97.38 \pm 1.18}$ & Kazakhstan & 99.92 & $86.33 \pm 3.06$ & $\mathbf{94.12 \pm 1.94}$\\
            Venezuala & 99.60 & $\mathbf{94.87 \pm 0.53}$ & $93.05 \pm 2.02$ & New Caledonia & 99.89 & $\mathbf{99.48 \pm 0.15}$ & $99.25 \pm 0.19$\\
            Philippines & 99.59 & $77.76 \pm 3.97$ & $\mathbf{81.95 \pm 8.07}$ & Estonia & 99.98 & $\mathbf{87.08 \pm 1.89}$ & $77.30 \pm 1.22$\\
            Slovenia & 99.88 & $95.37 \pm 0.23$ & $\mathbf{97.47 \pm 0.45}$ & Singapore & 99.46 & $86.42 \pm 2.36$ & $\mathbf{86.87 \pm 2.01}$\\
            Ukraine & 99.90 & $\mathbf{92.99 \pm 0.70}$ & $92.60 \pm 1.84$ & Bangladesh & 97.59 & $78.61 \pm 0.43$ & $\mathbf{82.45 \pm 2.54}$\\
            Belarus & 99.96 & $91.08 \pm 3.08$ & $\mathbf{96.40 \pm 1.76}$ & Paraguay & 99.82 & $96.01 \pm 1.23$ & $\mathbf{97.20 \pm 0.35}$\\
            Serbia & 99.76 & $\mathbf{95.31 \pm 0.48}$ & $92.62 \pm 3.83$ & Cyprus & 90.07 & $\mathbf{97.67 \pm 0.34}$ & $94.31 \pm 7.21$\\
            Croatia & 99.85 & $\mathbf{94.59 \pm 2.21}$ & $88.04 \pm 4.68$ & Bosnia & 99.62 & $84.04 \pm 1.47$ & $\mathbf{84.46 \pm 5.76}$\\
            Greece & 99.90 & $\mathbf{81.98 \pm 0.60}$ & $40.97 \pm 14.89$ & Ireland & 96.52 & $\mathbf{87.44 \pm 0.69}$ & $86.49 \pm 1.31$\\
            New Zealand & 99.73 & $94.27 \pm 1.50$ & $\mathbf{99.44 \pm 0.29}$ & Algeria & 98.86 & $\mathbf{85.37 \pm 2.05}$ & $84.65 \pm 4.47$\\
            Portugal & 99.59 & $\mathbf{93.65 \pm 0.46}$ & $92.68 \pm 1.46$ & Colombia & 97.70 & $87.81 \pm 0.92$ & $\mathbf{89.51 \pm 0.88}$\\
            Bulgaria & 99.24 & $91.03 \pm 2.07$ & $\mathbf{93.47 \pm 3.07}$ & Uzbekistan & 99.80 & $\mathbf{86.76 \pm 1.13}$ & $75.18 \pm 1.92$\\
            Lithuania & 99.99 & $\mathbf{87.67 \pm 3.05}$ & $76.41 \pm 1.66$ &  &  &  & \\
            \bottomrule
        \end{tabular}
\end{table*}

\begin{table*}[h!]
    \centering
    \caption{Zero-shot transfer models' z-test significance test per country(\textbf{bold} value are rejected null hypothesis with $\alpha = 0.001$)}
    \label{table:zero-shot-results-stats}
    \vspace{6pt}
    \begin{tabular}{cccccccc}
            \toprule
            Country &  \begin{tabular}[c]{@{}c@{}}Libpostal\\ FastText\end{tabular} & \begin{tabular}[c]{@{}c@{}}Libpostal\\ BPEmb\end{tabular} & \begin{tabular}[c]{@{}c@{}}FastText\\ BPEmb\end{tabular} & Country & \begin{tabular}[c]{@{}c@{}}Libpostal\\ FastText\end{tabular} & \begin{tabular}[c]{@{}c@{}}Libpostal\\ BPEmb\end{tabular} & \begin{tabular}[c]{@{}c@{}}FastText\\ BPEmb\end{tabular}\\
            \midrule
            Belgium & 88.71 & 92.22 & 4.70 & Réunion & -10.61 & -5.72 & 5.20\\
            Sweden & 80.64 & 55.54 & -33.76 & Moldova & 15.11 & 17.78 & 3.56\\
            Argentina & 59.55 & 54.68 & -6.27 & Indonesia & 29.42 & 26.00 & -4.28\\
            India & 86.04 & 62.05 & -28.71 & Bermuda & 10.42 & 8.95 & \textbf{-1.75}\\
            Romania & 31.88 & 40.10 & 11.04 & Malaysia & 19.05 & 6.29 & -14.26\\
            Slovakia & 60.15 & 42.70 & -23.21 & South Africa & 7.77 & 6.16 & \textbf{-2.12}\\
            Hungary & 108.57 & 143.25 & 45.25 & Latvia & 7.98 & 18.85 & 13.33\\
            Japan & 92.06 & 100.78 & 10.49 & Kazakhstan & 12.52 & 7.96 & -6.12\\
            Venezuala & 21.12 & 25.46 & 5.56 & New Caledonia & \textbf{1.63} & \textbf{2.22} & \textbf{0.68}\\
            Philippines & 49.84 & 44.09 & -7.57 & Estonia & 11.87 & 16.18 & 5.78\\
            Slovenia & 20.74 & 14.76 & -7.93 & Singapore & 11.19 & 10.97 & \textbf{-0.29}\\
            Ukraine & 25.79 & 26.54 & \textbf{1.03} & Bangladesh & 12.36 & 10.65 & \textbf{-2.05}\\
            Belarus & 26.44 & 16.41 & -13.53 & Paraguay & 5.46 & 4.43 & \textbf{-1.35}\\
            Serbia & 16.72 & 21.73 & 6.58 & Cyprus & -6.47 & \textbf{-3.23} & 3.50\\
            Croatia & 17.03 & 26.36 & 12.39 & Bosnia & 10.49 & 10.33 & \textbf{-0.22}\\
            Greece & 31.13 & 64.40 & 42.03 & Ireland & 5.98 & 6.43 & \textbf{0.50}\\
            New Zealand & 15.49 & \textbf{2.19} & -14.34 & Algeria & 8.68 & 8.95 & \textbf{0.35}\\
            Portugal & 15.84 & 17.27 & \textbf{1.85} & Colombia & 6.44 & 5.65 & \textbf{-0.91}\\
            Lithuania & 20.23 & 28.89 & 11.59 & Uzbekistan & 8.28 & 11.83 & 4.69\\
            Faroe Islands & 29.75 & 21.58 & -10.92 &  &  &  & \\
            \bottomrule
            \end{tabular}
\end{table*}

First, we observe that the \textbf{BPEmd} model reaches the highest accuracy most of the time. Indeed, $49~\%$ of the countries tested in zero-shot transfer reached a mean accuracy of at least $90~\%$ using \textbf{BPEmb} while using \textbf{fastText} only $46~\%$ of the countries reach that same accuracy. Most of these countries share either the same address structure or language proximity with training data. For instance, Venezuela shares the same address pattern as six other countries in the dataset and also shares the same language as Mexico, Spain, and the same Latin root as French. It was also interesting to observe that for Greece, \textbf{BPEmb} achieved near half the accuracy result as \textbf{fastText}. We hypothesize that for Greece fastText is able to produce better embeddings from subword units to reach this performance than \textbf{BPEmb}. 

In contrast, the lowest results (below $70~\%$) occur for countries where the address pattern and the country's official language were not seen in the training data such as India, Hungary, and Japan. The last two countries have had the lowest results of all. This is most likely due to the address structure (blue), which is the near inverse of the two most present ones (red and brown) (\autoref{fig:address_examples}). Indeed, since most of the other patterns used for the training are the opposite, our models were not exposed to this pattern during training. Also, those two countries do not share language root with any of the ones present in the training data, which makes the task difficult for our models. We also see that Kazakhstan, which uses the same address pattern as Japan, achieves better results. The main difference is the official language (Kazakh and Russian) presence in the training dataset. Moreover, India achieves almost $20~\%$ better results than Hungary and Japan, even if Hindi does not occur in the training dataset. This is probably due to the use of a nearly identical address pattern as the first one (red). The only difference being the inversion of the province and the postal code. It could mean that if no shared language root is present, a shared address structure allows a decent parsing of the address (almost $70~\%$). 

Second, we observe that both models (\textbf{fastText} and \textbf{BPEmb}) obtain decent results compared to Libpostal, considering that none of these countries were seen during training. Also, an interesting result is that of the Reunion zero-shot accuracy. The fact that this country shares the same address pattern as France (which was part of the training set) explains the good performance which has surpassed that of Libpostal despite not being trained directly on data from Reunion as opposed to Libpostal.

Finally, we observe that \textbf{BPEmp} achieves better results than \textbf{fastText} for 21 out of 41 countries, where most of them are between $1~\%$ to $20~\%$ better. Considering that nearly $81~\%$ of the countries reach an accuracy above $80~\%$, we conclude that using BPEmb embeddings gives good results for a zero-shot address parsing task considering that some languages and address patterns do not occur in the training data. To assess models' performance, we report the models' z-test significance test in \autoref{table:zero-shot-results-stats} using the same null hypothesis with $\alpha = 0.001$. We observe that \textbf{BPEmb} has a significantly better performance for 1 out of 41 countries compared to Libpostal, and we cannot reject the null hypothesis for 3 out of 41 countries, which means that performance between the two is not similar. Still, considering that the model was not trained, it's an interesting result. In addition, we observe that between \textbf{BPEmb} and \textbf{fastText} we have 12 out of 41 countries where the performance is similar and 15 out of 41 are in the advantage of \textbf{fastText}. These results were surprising to us since \textbf{fastText} uses pre-trained embedding trained on a French corpus. We hypothesize that fastText can produce better generalization of embeddings using subword units than \textbf{BPEmb}. This highlights that our trained model to combine the \textbf{BPEmb} embeddings might have overfitted to our problem due to the dataset size in contrast with fastText embeddings.

\section{Discussion}
We estimate that we have reached our first objective, which was to build a model capable of learning to parse addresses of different formats and languages using a multinational dataset and subword embeddings. As for our attempt at zero-shot transfer learning, it yielded interesting results. These results give us insights into the direction that our future work should take. We would also like to state that due to the time and resources consuming nature of the training process, we have not been able to perform a grid search to find optimal hyperparameters for our models. It could be interesting to explore how other subword embeddings techniques, such as the character-based ones, would perform on the multinational address parsing task. Adding an attention mechanism \cite{bahdanau2014neural} could also give interesting insights into the address elements on which the model focuses when making a tag prediction.

\section{Conclusion}

In this article, we tackled the multinational address parsing problem and proposed a solution based on the use of subword embeddings to solve the multilingual aspect of the problem, as well as a sequence-to-sequence model for the tagging task. Our approach was able to reach state-of-the-art results on address from twenty countries and despite the different address formatting systems without the use of any pre-processing nor post-processing. We have also explored the possibility of zero-shot transfer across countries and achieved interesting, but not yet optimal results. As part of our future work, we will aim at applying domain adaptation techniques to better transfer the learned knowledge about a country's address parsing to other countries' addresses.

\section*{Acknowledgment}
This research was supported by the Natural Sciences and Engineering Research Council of Canada (IRCPJ 529529-17) and a Canadian insurance company. We wish to thank the reviewers for their comments regarding our work and methodology.

%\appendix

%\subsection{Training and Validation Loss of \textbf{BPEmb}}
%\label{app:loss}
%As discussed in \autoref{sec:exp}, we have trained of our two models five-time using a different seed each time. In this complementary section, we discuss why we chose to train a sixth \textbf{BPEmb} model using the seed 30. The \autoref{fig:loss} present the training and validation loss of those five trained models for \textbf{BPEmb}. We observe that one obtains (purple) poor results among the training loss (left) compared to the others. The same results can be seen on the validation loss (right) obtained during training. These results highlight that the training of the fifth model did not converge. Since the difference between this model's final loss (around 4) is quite higher than those obtained by the four other models (below 0.1), we chose to train a sixth model using the seed 30 and use the first four and the sixth to generate our aforementioned evaluation results. 

%\begin{figure}[h!]
%    \centering
%    \includegraphics[scale=0.45, keepaspectratio]{ressources/images/loss_graph.png}
%    \caption{Training and validation loss of all five \textbf{BPEmb} model}
%    \label{fig:loss}
%\end{figure}

\bibliographystyle{IEEEtran}
\bibliography{MNLP2020}

\vspace{12pt}

\end{document}